\documentclass[10pt,twocolumn,letterpaper]{article}

\usepackage{wacv}
\usepackage{times}
\usepackage{epsfig}
\usepackage{graphicx}
\usepackage{amsmath}
\usepackage{amssymb}

\usepackage{tabularx}
\usepackage{bm}

\wacvfinalcopy 

\def\wacvPaperID{405} 

\ifwacvfinal\fi
\begin{document}

\title{Improving Image Captioning by Leveraging Knowledge Graphs}


\author{Yimin Zhou, Yiwei Sun, Vasant Honavar\\
Artificial Intelligent Research Laboratory\\
The Pennsylvania State University\\
{\tt\small zhoumingdetiger@gmail.com, yus162@psu.edu, vhonavar@ist.psu.edu}
}

\maketitle
\ifwacvfinal\thispagestyle{empty}\fi

\begin{abstract}
We explore the use of a knowledge graphs, that capture general or commonsense knowledge, to augment the information extracted from images by the state-of-the-art methods for image captioning. The results of our experiments, on several benchmark data sets such as MS COCO, as measured by CIDEr-D,  a performance metric  for image captioning, show that the variants of the state-of-the-art methods for image captioning that make use of the information extracted from knowledge graphs can substantially outperform those that rely solely on the information extracted from images.

\end{abstract}


\section{Introduction}
Advances in digital technologies have made it possible to acquire and share vast amounts of data of all kinds, including in particular, images. The availability of such data, together with recent advances in machine learning, has resulted in robust and practical machine learning based solutions to object recognition, e.g., Inception\cite{szegedy2015going}, vgg16\cite{DBLP:journals/corr/SimonyanZ14a}, ResNet\cite{he2016deep}.

Recent years have witnessed a growing interest in describing visual scenes, a task that is remarkably easy for humans yet remains difficult for machines \cite{fei2007we}. Of particular interest in this context is the image captioning problem, which requires analyzing the visual content of an image, and generating a caption, i.e., a textual description that summarizes the most salient aspects of the image. Just as question answering presents challenges beyond text processing, image captioning presents several challenges beyond image processing. Effective image captions need to provide information that is not explicit in the image, e.g., "People gathered to watch a volleyball match" when describing a crowd seated around a volleyball court, even if the image shows no players on the field (perhaps because the game is yet to begin), or "An impressionist painting of a garden by Claude Monet", even if the image makes no explicit mention of Monet or impressionism.  Generating such captions calls for incorporating background knowledge with information that is available in the image.  However, existing methods for image captioning (See  \cite{bernardi2016automatic} for a review) fail to take advantage of readily available general or commonsense knowledge about the world, e.g., in the form of {\em knowledge graphs}.   

Inspired by the success of information retrieval and question answering systems that leverage background knowledge  \cite{wang2017knowledge}, we explore an approach to image captioning that uses information encoded in knowledge graphs. Specifically, we augment the neural image caption (NIC) method introduced in \cite{vinyals2015show,you2016image} where a convolutional neural network (CNN) \cite{krizhevsky2012imagenet} trained to encodes an image into a fixed length vector space representation or embedding and uses the embedding to specify the initial state of a recurrent neural network (RNN) that is trained to produce sentences describing the image in two important aspects: In addition to a CNN trained to generate vector space embedding of image features, we use an object recognition module that given an image as input, produces as output, a collection of terms that correspond to objects in the scene We use an external knowledge graph, specifically, ConceptNet \cite{liu2004conceptnet,speer2017conceptnet}, a labeled graph which connects words and phrases of natural language connected by edges that denote {\em commonsense} relationships between them, to infer a set of terms directly or indirectly related to the words that describe the objects found in the scene by the object recognition module. Vector space embeddings of the terms as well as the image features are then used to specify the initial state of an LSTM-based RNN that is trained to produce the caption for the input image. We call the resulting image captioning system ConceptNet enhanced neural image captioning system (CNet-NIC).  The results of our experiments on the MS COCO captions benchmark dataset \cite{lin2014microsoft} show that CNet-NIC is competitive with or outperforms the state-of-the-art image captioning systems on several of the commonly used performance measures (BLEU \cite{papineni2002bleu} , METEOR\cite{banerjee2005meteor}, ROUGE-L\cite{lin2004rouge}, all of which are measures designed originally for evaluating machine translation systems as opposed to image captioning systems). More importantly, CNet-NIC substantially outperforming the competing methods on CIDEr-D, a variant of the CIDEr  \cite{vedantam2015cider}, the only measure that is designed explicitly for evaluating image captioning systems. Because CIDEr-D measures the similarity of a candidate image caption  to a collection of  human generated reference captions, our results suggest that the incorporation of background knowledge from ConceptNet enables CNet-NIC to produce captions that are more similar to those generated by humans than those produced by  methods that do not leverage such background knowledge.

The rest of the paper is organized as follows. Section 2 summarizes the related work on image captioning that sets the stage for our work on CNet-NIC. Section 3 the design and implementation of CNet-NIC. Section 4 describes our experimental setup and the results of our experiments assessing the performance of CNet-NIC on the MS COCO image captioning benchmark dataset along with comparisons with the competing state-of-the-art methods using the standard performance measures (BLEU@$N$ ($N\in{1,2,3,4}$), METEOR, ROUGE-L, and CIDEr-D) as well as a qualitative analysis of a representative sample of the captions produced by CNet-NIC. Section 5 concludes with a summary and an outline of some directions for further research.

\section{Related Work}

Existing image captioning methods can be broadly grouped into the following (not necessarily disjoint) categories: (i) Template-based methods e.g., \cite{barnard2003matching,gupta2012image,farhadi2010every,kulkarni2013babytalk} which rely on (often hand-coded) templates. Such methods typically detect the object types, their attributes, scene types (e.g., indoor versus outdoor), etc., based on a set of visual features, and generate image captions by populating a template with the information extracted from the image. (ii) Retrieval-based methods which can be further subdivided into two groups: (ii.a) Image similarity based methods e.g., \cite{ordonez2011im2text,gupta2012image,kuznetsova2012collective,devlin2015exploring,kuznetsova2014treetalk} 
which retrieve captioned images that are visually most similar to the target image and transfer their captions to the target image; and (ii.b) Multimodal similarity based methods that use features of images as well as the associated captions to retrieve or synthesize the caption for the target image \cite{hodosh2013framing,frome2013devise,socher2014grounded,karpathy2014deep,jia2015guiding,ma2015multimodal,vinyals2015show}; (iii) Embedding-based methods, including those that use recurrent, convolutional, or deep neural networks  \cite{vinyals2015show,xu2015show,wu2016value,you2016image,DBLP:journals/corr/MaoXYWY14,yao2016boosting,vinyals2017show,donahue2015long} that make use of the learned low-dimensional embeddings of images to train caption generators.

However, none of the existing methods take advantage of the readily available background knowledge about the world e.g., in the form of {\em knowledge graphs}. Such background knowledge has  been shown to be useful in a broad range of applications ranging from information retrieval to question answering \cite{wang2017knowledge}, including most recently, visual question answering (VQA) from images \cite{wu2017image}. We hypothesize that such background knowledge can address an important drawback of existing image captioning methods, by enriching captions with information that is not explicit in the image. 

Unlike the state-of-the-art image captioning systems, CNet-NIC is specifically designed to take advantage of background knowledge to augment the information extracted from the image (image features, objects) to improve machine-produced captions or image descriptions. Unlike VQA  \cite{wu2017image}, which uses a knowledge graph to extract better image features and hence better answer questions about the image, CNet-NIC first detects objects (not just image features) in the image and uses the detected objects to identify related terms or concepts which are then used to produce better image captions.

\section{CNet-NIC: ConceptNet-Enhanced Neural Image Captioning}
We proceed to describe our design for an image captioning system that takes advantage of background knowledge in the form of a knowledge graph. 

\subsection{CNet-NIC Architecture}

\begin{figure*}
\centering
\includegraphics[width=6.2in]{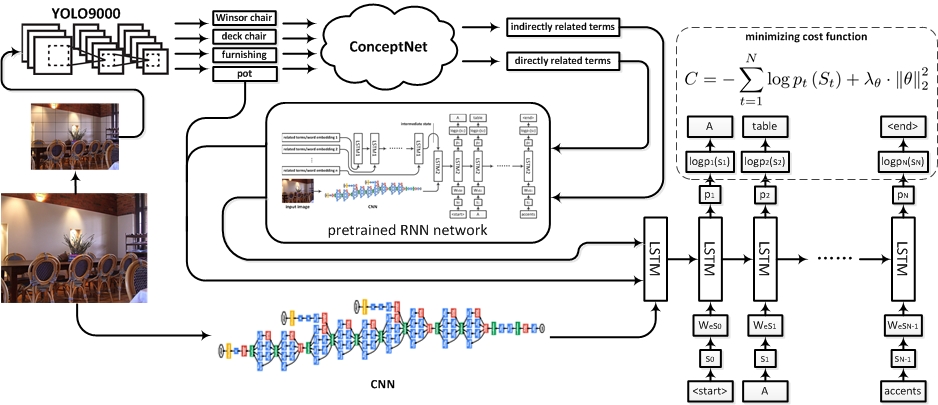}
\caption{Our model architecture by inviting common sense from external resources}
\label{fig_ourmodelarchitecture}
\end{figure*}

Fig. \ref{fig_ourmodelarchitecture} shows a schematic of the CNet-NIC system. CNet-NIC uses YOLO9000\cite{redmon2017yolo9000}, a state-of-the-art general-purpose real-time object recognition module that is trained to recognize 9000 object categories. YOLO9000 takes an image as input and produces as output, a collection of terms that refer to objects in the scene. CNet-NIC use an external knowledge graph, specifically, ConceptNet \cite{liu2004conceptnet,speer2017conceptnet}, a labeled graph which connects words and phrases of natural language connected by edges that denote {\em commonsense} relationships between them, to infer two sets  of terms related to the words that describe the objects found in the scene by the object recognition module. The first set of terms are retrieved based on the individual objects in the scene. The second set of terms are retrieved based on the entire collection of objects in the scene. The resulting terms are then provided to a pre-trained RNN to obtain the corresponding vector space embedding of the terms. A CNN is used to obtain vector space embedding of the image features.  The two resulting vector space embeddings are  used to specify the initial state of an LSTM-based RNN which is trained to produce the caption for the input image. 

We proceed to describe each key element of the CNet-NIC system in detail.

\subsection{Improving Image Captioning by Incorporating Background Knowledge}
To test our hypothesis, we use the ConceptNet\cite{liu2004conceptnet,speer2017conceptnet}, a kind of knowledge graph, specifically, one that connects words and phrases of natural language connected by edges that denote {\em commonsense} relationships between them. ConceptNet integrates information from resources provided by experts as well as through crowd-sourcing. It encodes general knowledge that is of use in natural language understanding, and has been shown to enrich the semantic information associated with words, beyond that supplied by distributional semantics \cite{speer2017conceptnet}.

\subsection{Generating Semantic Representations from ConceptNet}
ConceptNet can be used to learn word embeddings using a variant of  "retrofitting" \cite{faruqui2015retrofitting}. Let $V=\left\{w_1,\dots,w_n\right\}$ be a vocabulary, i.e., the set of word types, and $\Omega$ be an ontology encoding semantic relations between words in $V$. $\Omega$ is represented as an undirected graph $(V,E)$ with one vertex for each word type and edges $(w_i,\;w_j)\in E\subseteq V\times V$ indicating a semantic relationship of interest.

Let $\hat Q$ be the collection of vectors $\hat{q_i} \in \mathbb{R}^{d}$ for each $w_i \in V$ that is learned using a standard data-driven method where $d$ is the length of word vectors. The objective is to learn the matrix $Q=(q_1, \dots, q_n)$ such that the columns are both close to their counterparts in $\hat Q$ and to adjacent vertices in $\Omega$ where closeness is measured using an appropriate distance measure, e.g., the Euclidean distance. This is achieved by minimizing the following objective function:

\begin{equation}
\Psi\left(Q\right)=\sum_{i=1}^n\left[\alpha_i\left\|q_i-\hat{q_i}\right\|^2+\sum_{\left(i,j\right)\in E}\beta_{ij}\left\|q_i-q_j\right\|^2\right]
\end{equation}

where $\alpha$ and $\beta$ are  parameters that control the the relative strengths of associations. The procedure is called retro-fitting because the word vectors are first trained independent of the information in the semantic lexicons and  are then retro-fitted by optimizing the objective function specified above. Because $\Psi$ is convex in $Q$, the solution of the resulting optimization problem is straightforward. $Q$ can be initialized to $\hat Q$ and iteratively updated using the following update equation:

\begin{equation}
q_i=\frac{\sum_{j:\left(i,j\right)\in E}\beta_{ij}q_j+\alpha_i\hat{q_i}}{\sum_{j:\left(i,j\right)\in E}\beta_{ij}+\alpha_i}
\end{equation}

\subsection{Simple Recurrent Neural Network Image Caption Generator}

\begin{figure*}[h]
\centering
\includegraphics[width=3in]{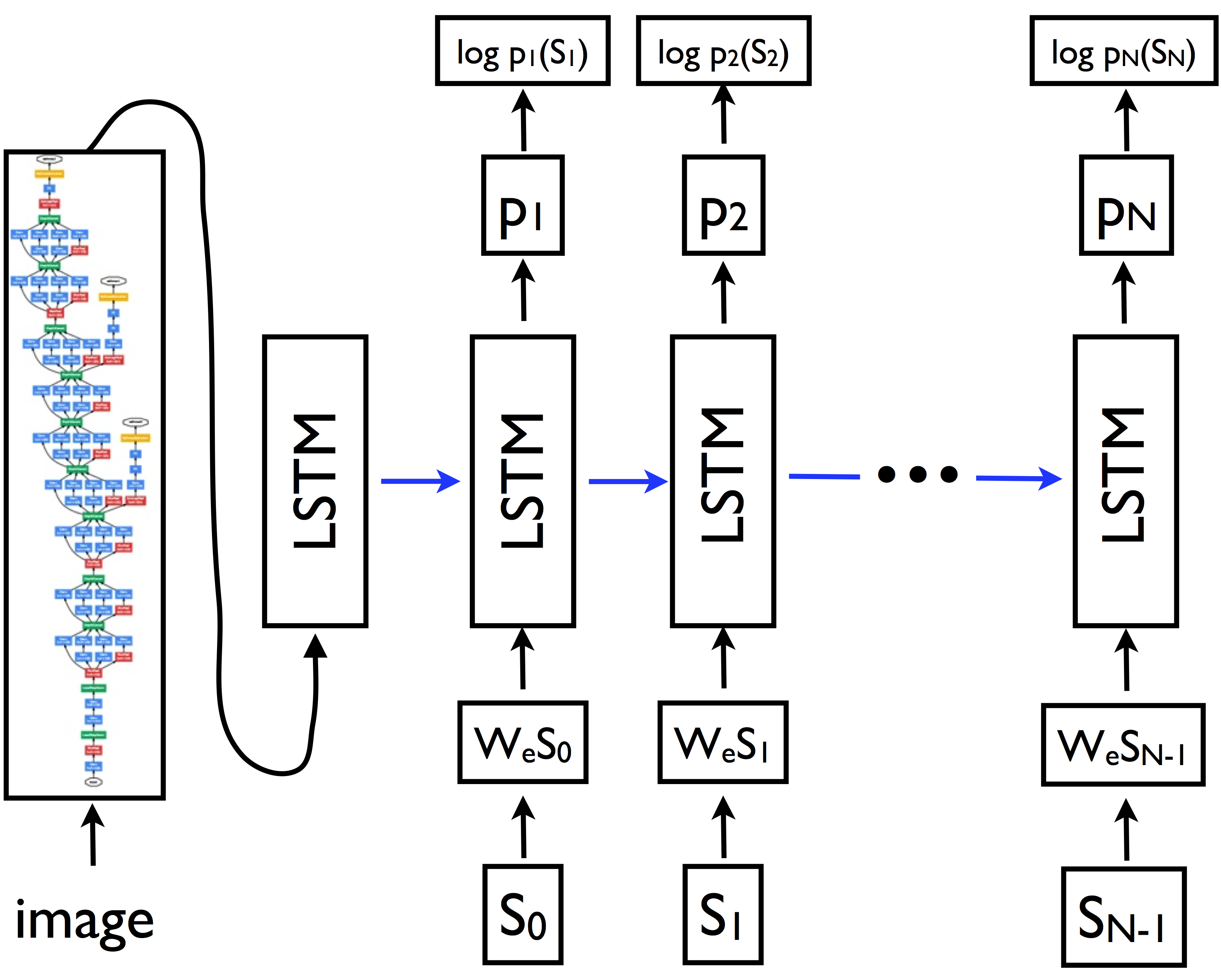}
\caption{The architecture of the simple recurrent neural network image caption generator}
\label{fig_architectureshowandtell}
\end{figure*}

We use a simple recurrent neural network image caption generator based on LSTM introduced in \cite{vinyals2015show} where a CNN is used to extract image features; and  vector space embedding of the extracted features is used by an LSTM-based RNN to generate the caption text. The architecture of this model is shown in Fig. \ref{fig_architectureshowandtell}. 

Let $X$ be an input image and  $S=\left(S_0, \dots ,S_N\right)$ the corresponding caption sentence.  Let

\begin{equation}
x_{-1}\;=\;CNN\left(X\right)
\end{equation}

\begin{equation}
x_t\;=\;W_eS_t,\;\;\;t\in\left\{0\;\dots\;N-1\right\}
\end{equation}

where $S_t$ is the one-hot vector representation of the word with a size of the dictionary, $S_0$ a special start word, and $S_N$ a special end word.

\begin{equation}
p_{t+1}\;=\;LSTM\left(x_t\right),\;\;\;t\in\left\{0\;\dots\;N-1\right\}
\end{equation}
 
The loss function is given by:

\raggedbottom

\begin{equation}
L\left(I,S\right)=-\sum_{t=1}^N\log p_t\left(S_t\right)
\end{equation}

The loss function is minimized with respect to the parameters of the LSTM, CNN and $W_e$.

\subsection{Identifying Semantically Related Words}

Given  set of input words $W=\{w_1\cdots w_n\}$,  and their associated weights $u_1\cdots u_n$ (e.g., based on their frequency distribution), a target word $w$ can be scored based on its semantic relatedness to the input words, as measured by the weighted distance (e.g., cosine distance) between the semantic vector representation of the query word with each of the target words. Let $s_w$ and $s_{w_i}$ denote the semantic vector representations of words $w$ and $w_i$ ($i\in \{1,\cdots n\}$; and $d(a,b)$ denote the (cosine) distance between vectors $a$ and $b$.

\begin{equation}
score(W,w)=\frac{{\displaystyle\sum_{i=1}^n}u_i d(s_w,s_{w_i})}{\sum_{i=1}^nu_i}
\end{equation}

\begin{figure*}[htp]
\centering
\includegraphics[width=4.5in]{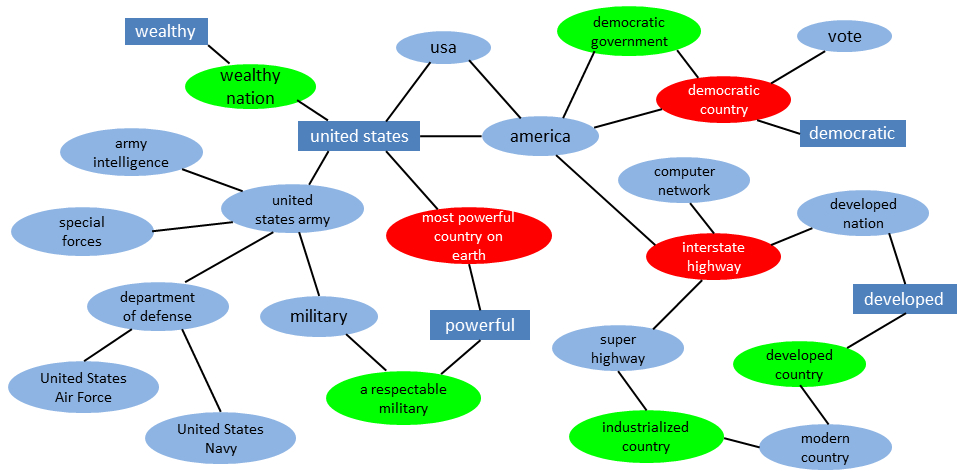}
\caption{An example of relevant words generated by ConceptNet}
\label{fig_conceptnetsemanticweb}
\end{figure*}

Fig. \ref{fig_conceptnetsemanticweb}  Identifying Semantically Related Words.  Blue rectangular nodes denote the input words (concepts). Red ovals denote the words that are most closely related to the input words, whereas the green ovals the next most closely related, and light blue ovals the next most closely related.

\subsection{CNet-NIC}
Let $X$ be an input image, and $O$ a set of terms corresponding to the objects detected in the image $I$ by the YOLO9000 object recognition system. Thus, $O=YOLO\left(X\right)$. For each $o\in O$, let
$r_o=ConceptNet\left(o\right)$ be the set of terms related to $o$ in the ConceptNet knowledge graph; and  
$R_O=ConceptNet\left(O\right)$ the set of terms related to the entire set $O$ of terms referring to all of the objects detected in the image $X$ by the YOLO9000 object recognition system. Let $D=
\bigcup_{o\in O}r_o\cup\left\{o\right\}$ denote the set of terms directly related to individual objects in $X$. 
Loosely speaking, $R_O$ provides terms that are descriptive of  the scene as a whole, whereas $I$ provides terms that are descriptive of some or all of the objects depicted in the image. Thus, $I=R_O-D$ denote a set of terms that are indirectly related to objects in $X$. Let $
d=RNN_D\left(D\right)$ and $i=RNN_I\left(I\right)$ denote the vector space embeddings of $D$ and $I$ produced by the pre-trained RNNs $RNN_D$ and $RNN_I$ respectively. Let $a=CNN\left(X\right)$ be an embedding of the image features of $X$ produced by a pre-trained CNN. The image captions are produced by an LSTM-based RNN whose state is initialized as follows:

\begin{equation}
x_{-1}\;=\;a \parallel d \parallel i
\end{equation}

where $\parallel$ denotes the concatenation operation. 

\begin{equation}
x_t\;=\;W_eS_t,\;\;\;t\in\left\{0\;\dots\;N-1\right\}
\end{equation}

where $S_t$ denotes the one-hot vector representation of the word with a size of the dictionary, and $S_0$ a special start word, and $S_N$ a special end word.

\begin{equation}
p_{t+1}\;=\;LSTM\left(x_t\right),\;\;\;t\in\left\{0\;\dots\;N-1\right\}
\end{equation}

 The cost function is given by:

\begin{equation}
C=-\sum_{t=1}^N\log p_t\left(S_t\right)+\lambda_\theta\cdot\left\|\theta\right\|_2^2
\end{equation}

where $\theta$ represents the model parameters and $\lambda_\theta\cdot\left\|\theta\right\|_2^2$ is a regularization term.

\begin{figure*}
\centering
\includegraphics[width=4.2in]{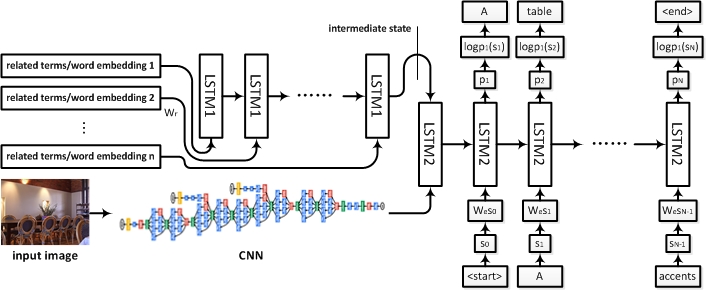}
\caption{The pre-trained RNN network architecture}
\label{fig_pretrainedrnnmodelarchitecture}
\end{figure*}

The only trainable parameters are within the LSTM and $W_e$. The pre-trained RNN network is shown in Fig. \ref{fig_pretrainedrnnmodelarchitecture}. Let $r_i$ be the $i$th word embedding.

\begin{equation}
x_i=W_rr_i,\;\;\;i\in\{i,\;\dots\;L-1\}
\end{equation}

\begin{equation}
a=CNN\left(X\right)
\end{equation}

$a$ represents the image attribute embedding from CNN.

\begin{equation}
y_{-1}=concatenate\left(a,\;x_{L-1}\right)
\end{equation}

\begin{equation}
a=CNN\left(X\right)
\end{equation}

\begin{equation}
y_t=W_eS_t,\;\;\;t\in\left\{0,\;...\;N-1\right\}
\end{equation}

\begin{equation}
p_{t+1}\;=\;LSTM\left(x_t\right),\;\;\;t\in\left\{0\;\dots\;N-1\right\}
\end{equation}

The cost function to be minimized is given by:

\begin{equation}
C=-\sum_{t=1}^N\log p_t\left(S_t\right)
\end{equation}

Note that the only  parameters to be trained are those associated with LSTM1, LSTM2 and $W_e$. The trained LSTM1 is used in the model in Fig. 3 to extract intermediate state of the related terms.

\begin{table*}[htp]

\footnotesize
\caption{Performance of our proposed models and other state-of-the-art methods on MS COCO dataset, where B@$N$, M, R, and C are short for BLEU@$N$, METEOR, ROUGE-L, and CIDEr-D scores. Except CIDEr-D, all values are reported as percentage(\%).}

\begin{center}
\label{table_performancecomparison}
\begin{tabularx}{4.225in}{ c | c | c | c | c | c | c | c }
\hline
\textbf{Model}  &  \textbf{B@1}  &  \textbf{B@2}  &  \textbf{B@3}  &  \textbf{B@4}  &  \textbf{M}  &  \textbf{R}  &  \textbf{C}  \\  \hline
\textbf{NIC}\cite{vinyals2015show}  &  $66.6$  &  $45.1$  &  $30.4$  &  $20.3$  &  -  &  -  &  -  \\
\textbf{LRCN}\cite{donahue2015long}  &  $62.8$  &  $44.2$  &  $30.4$  &  $21$  &  -  &  -  &  -  \\
\textbf{Soft Attention}\cite{xu2015show}  &  $70.7$  &  $49.2$  &  $34.4$  &  $24.3$  &  $23.9$  &  -  &  -  \\
\textbf{Hard Attention}\cite{xu2015show}  &  $71.8$  &  $50.4$  &  $35.7$  &  $25$  &  $23$  &  -  &  -  \\
\textbf{ATT}\cite{you2016image}  &  $70.9$  &  $53.7$  &  $40.2$  &  \textit{30.4}  &  $24.3$  &  -  &  -  \\
\textbf{Sentence Condition}\cite{DBLP:journals/corr/ZhouXKC16}  &  $72$  &  $54.6$  &  $40.4$  &  $29.8$  &  $24.5$  &  -  &  $95.9$  \\
\textbf{LSTM-A}\cite{yao2017boosting}  &  \textit{73}  &  $\bm{56.5}$  &  $\bm{42.9}$  &  $\bm{32.5}$  &  \textit{25.1}  &  \textit{53.8}  &  \textit{98.6}  \\ \hline
\textbf{CNet-NIC}  &  $\bm{73.1}$  &  \textit{54.9}  &  \textit{40.5}  &  $29.9$  &  $\bm{25.6}$  &  $\bm{53.9}$  &  $\bm{107.2}$  \\
\hline

\end{tabularx}
\end{center}
\end{table*}

\begin{table*}[htp]

\footnotesize

\caption{Performance of variants of CNet-NIC on MS COCO dataset, where B@$N$, M, R, and C are short for BLEU@$N$, METEOR, ROUGE-L, and CIDEr-D scores. Except CIDEr-D, all values are reported as percentage(\%).}

\begin{center}
\label{table_factoranalysis}
\begin{tabularx}{5.725in}{ p{7cm} | c | c | c | c | c | c | c }
\hline
\textbf{Input of Model}  &  \textbf{B@1}  &  \textbf{B@2}  &  \textbf{B@3}  &  \textbf{B@4}  &  \textbf{M}  &  \textbf{R}  &  \textbf{C}  \\  \hline
\textbf{none(only seqs input)}  &  $48.4$  &  $24.7$  &  $10.2$  &  $3.9$  &  $11$  &  $34.2$  &  $8.6$  \\  \hline
\textbf{image embedding}  &  $70.3$  &  $52.9$  &  $38.3$  &  $27.5$  &  $24.3$  &  $51.8$  &  $99.5$  \\  \hline
\textbf{detected objects and directly related terms}  &  $63.3$  &  $43.4$  &  $29.1$  &  $20$  &  $19.8$  &  $46.1$  &  $74.3$  \\  \hline
\textbf{indirectly related terms}  &  $47.6$  &  $27$  &  $15.7$  &  $10.2$  &  $13.7$  &  $36.6$  &  $31.8$  \\  \hline
\textbf{detected objects and directly related terms + image embedding}  &  $70.9$  &  $53.3$  &  $38.7$  &  $28$  &  $24.8$  &  $52.4$  &  $103.2$  \\  \hline
\textbf{indirectly related terms + image embedding}  &  $70.1$  &  $52.8$  &  $38.2$  &  $27.7$  &  $24.5$  &  $52$  &  $100.5$  \\  \hline
\textbf{detected objects and directly related terms + indirectly related terms + image embedding}  &  $72.1$  &  $54.2$  &  $38.9$  &  $28.5$  &  $24.8$  &  $52.9$  &  $103.6$  \\  \hline
\textbf{detected objects and directly related terms + indirectly related terms + image embedding + fine tune CNN}  &  $73.1$  &  $54.7$  &  $40.5$  &  $29.9$  &  $25.6$  &  $53.9$  &  $107.2$  \\

\hline

\end{tabularx}
\end{center}
\end{table*}

\section{Experiments}
\subsection{Data}
We used the Microsoft COCO captioning data set(COCO)\cite{lin2014microsoft}, the most widely used image captioning benchmark data set in our evaluations. The data set includes 82,783 images for training and 40,504 images for validation. For each image, the data set includes  5 or 6 descriptions or captions provided by human annotators. In our experiments, from the training and validation set provided, we used  117,211 images for training, 2,026 images for validation and 4050 images for testing.

\subsection{Experimental Setup}
Details of the experimental setup are summarized below:
\begin{itemize}
\item {\bf Data Preprocessing}: Following \cite{karpathy2014deep}, we convert all of the image captions in training set to lower case and discard rare words which occur less than 4 times, resulting in the final vocabulary with 11,519 unique words in COCO data set. Each word in the sentence is represented as "one-hot" vector. 

\item {\bf Attribute Extraction}: To extract the image attributes, we use the output of inception v3\cite{szegedy2016rethinking} network image recognition model pre-trained on the ILSVRC-2012-CLS\cite{everingham2010pascal} image classification data set.

\item {\bf Object Detection}: We use the YOLO9000 object detection network (with $23$ layers) and $544\times544$ resolution. YOLO9000 is able to detect 9419 object classes.

\item {\bf Leveraging Background Knowledge}:  Because YOLO9000 object detection system is inherently imperfect, in identifying related terms using ConceptNet, we limit ourselves to only the objects detected with high confidence. Based on preliminary experiments, we set $30\%$ as the detection threshold.

\item {\bf Training the model}: Our model is implemented on the TensorFlow platform in Python language. The size of LSTM for each embedding(attributes, related terms) is set to $512$. Initial learning rate is set to $2.0$  with an exponential decay schedule. Batch size is set to $32$. Along the training, the learning rate is shrunk by $5$ for three or four times. The number of iterations is set to 500,000.

\item {\bf Testing the model}: Two approaches can be utilized for sentence generation during the testing stage. One approach is to select the word with maximum probability at each time step and set it as LSTM input for next time step until the end sign word is emitted or the maximum length of sentence is reached. Another approach is to conduct a beam search that selects the top-$k$ best sentences at each time step and use them as the candidates to generate the top-$k$ best sentences at the next time step. We adopt the second approach and set the beam size $k$ empirically to $3$.

\item {\bf Evaluation Metrics}: To evaluate CNet-NIC, we use 4 metrics: BLEU@$N$\cite{papineni2002bleu}, METEOR\cite{banerjee2005meteor}, ROUGE-E\cite{lin2004rouge}, and CIDEr-D\cite{vedantam2015cider}. All the metrics are computed by using the codes released by \cite{DBLP:journals/corr/ChenFLVGDZ15}.
\end{itemize}

\begin{table*}[htp]
\scriptsize
\caption{Image example of model showing "Common Sense" from external resource.}

\begin{center}
\label{table_invitingcommonsense}

\resizebox{16cm}{9cm}{

\begin{tabular}{ c | p{1.7cm} | p{1.7cm} | p{2.9cm} | p{2.9cm} | p{2.9cm} }

\hline
\textbf{Image}  &  \textbf{Detected}  &  \textbf{Indirectly Related}  &  \textbf{Sentences Generated by Model with Indirectly Related}  &  \textbf{Sentences Generated by Model without Indirectly Related}  &  \textbf{Standard Model}  \\  \hline
\raisebox{-1.15\height}{\includegraphics[width=0.52in]{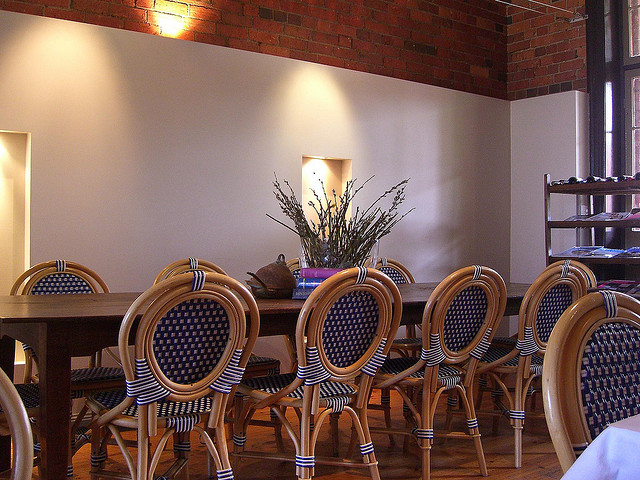}}  &  Winsor chair, deck chair, furnishing, pot  &  item of furniture, \textbf{upholstered}, \textbf{found in house}, \textbf{chairs}  &  0) a dining room with a \textbf{table} and \textbf{chairs}\newline 1) a dining room with a table, chairs and a table\newline 2) a dining room with a table and chairs and a \textbf{fireplace}  &  0) a table with a vase of flowers on it\newline 1) a dining room with a table and chairs\newline 2) a table with a vase of flowers on it  &  0) a table with a vase of flowers on it\newline 1) a table with a vase of flowers on it\newline 2) a table with a vase of flowers on it  \\  \hline
\raisebox{-1.4125\height}{\includegraphics[width=0.52in]{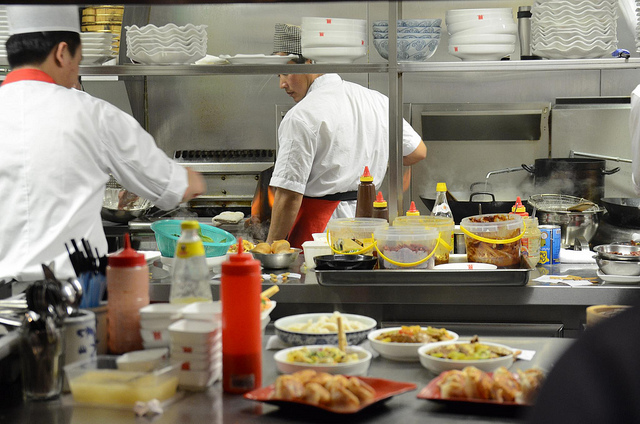}}  &  fishmonger, cereal bowl, phial, banana, waiter  &  \textbf{food storage jar}, canaree, \textbf{storing food}, fruit bowl, food can  &  0) a \textbf{chef} preparing food in a kitchen on a counter\newline 1) a \textbf{chef} preparing food in a kitchen on a table\newline 2) a man in a kitchen preparing food for a customer  &  0) a man and a woman preparing food in a kitchen\newline 1) a man and a woman preparing food in a kitchen\newline 2) a chef preparing food in a kitchen next to a woman  &  0) a group of people in a kitchen preparing food\newline 1) a group of people standing around a kitchen\newline preparing food
2) a group of people in a kitchen preparing food  \\  \hline
\raisebox{-0.9125\height}{\includegraphics[width=0.50in]{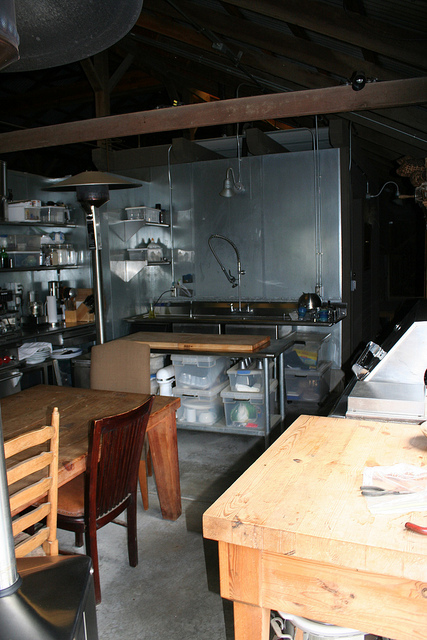}}  &  straight chair, furnishing  &  \textbf{item of furniture}, \textbf{reupholstery}, end table  &  0) a kitchen \textbf{filled with appliances and lots of clutter}\newline 1) a kitchen \textbf{filled with appliances and lots of counter space}\newline 2) a kitchen {with a table and chairs}  &  0) a kitchen with a stove a sink and a counter\newline 1) a kitchen with a stove a sink and a window\newline 2) a kitchen with a stove top oven next to a sink  &  0) a kitchen with a stove a sink and a stove\newline 1) a kitchen with a stove a sink and a refrigerator\newline 2) a kitchen with a stove a sink and a counter  \\  \hline
\raisebox{-0.875\height}{\includegraphics[width=0.50in]{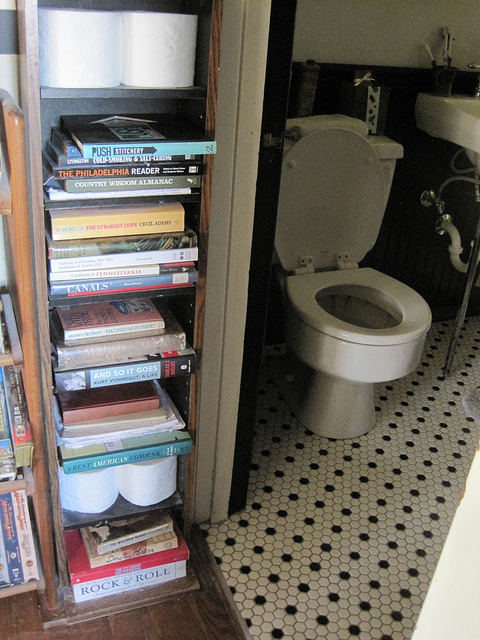}}  &  book(s), toilet seat  &  \textbf{bookrack}, \textbf{bookshelving}, \textbf{bookrest}  &  0) a bathroom with a toilet and a \textbf{book shelf}\newline 1) a bathroom with a toilet and a book shelf\newline 2) a bathroom with a toilet and a sink  &  0) a white toilet sitting in a bathroom next to a wall\newline 1) a white toilet sitting next to a book shelf\newline 2) a white toilet sitting in a bathroom next to a shelf  &  0) a kitchen with a stove a sink and a stove\newline 1) a kitchen with a stove a sink and a refrigerator\newline 2) a kitchen with a stove a sink and a counter  \\  \hline
\raisebox{-1.30\height}{\includegraphics[width=0.52in]{COCO_train2014_000000215481.jpg}}  &  trolleybus(es), park bench, commuter  &  \textbf{tram stop}, bus rapid transit  &  0) a couple of buses that are \textbf{sitting} in the street\newline 1) a couple of buses that are \textbf{parked next to each other}\newline 2) a couple of buses driving down a street next to a tall building  &  0) a double decker bus driving down a street\newline 1) a double decker bus driving down the street\newline 2) a double decker bus is driving down the street  &  0) a double decker bus driving down a street\newline 1) a double decker bus driving down a city street\newline 2) a double decker bus driving down the street  \\  \hline
\raisebox{-1.160\height}{\includegraphics[width=0.52in]{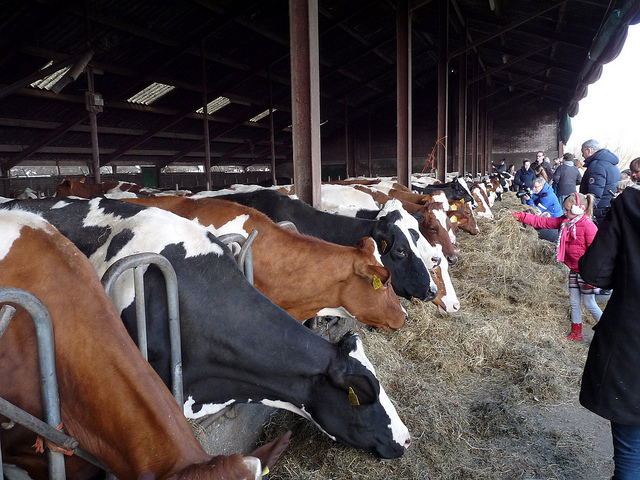}}  &  Friesian(s), Brown Swiss, private, settler  &  \textbf{dairy farm}, \textbf{cows}, \textbf{feed lot}  &  0) a group of cows standing next to each other\newline 1) a group of cows that are standing in the dirt\newline 2) a group of cows standing in a \textbf{barn}  &  0) a group of cows that are standing in the dirt\newline 1) a group of cows that are standing in the grass\newline 2) a group of cows that are standing in a pen  &  0) a group of cows are standing in a pen\newline 1) a group of cows standing in a pen\newline 2) a group of cows are standing in a field  \\  \hline

\raisebox{-0.8875\height}{\includegraphics[width=0.50in]{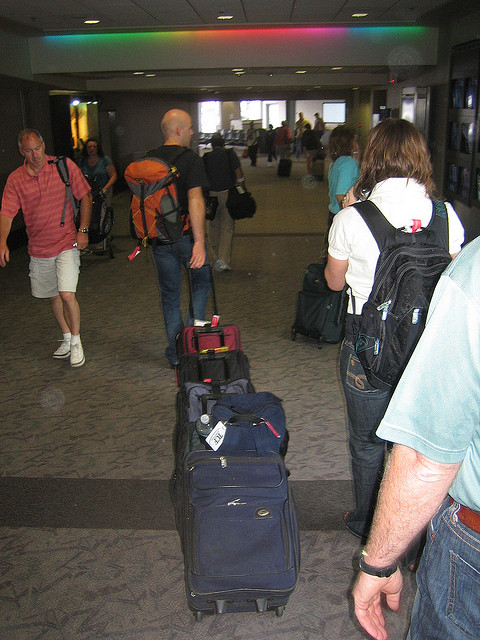}}  &  overnighter(s), pilgrim(s), square dancer, general, peddler, hand luggage, backpack  &  \textbf{wayfaring}, \textbf{day tripper}, \textbf{journeyer}, \textbf{excursionist}, \textbf{traveller}  &  0) a group of people standing around with luggage.\newline 1) a group of people standing around a luggage carousel.\newline 2) a group of standing around a \textbf{baggage claim area}.  &  0) a group of people standing around a luggage carousel.\newline 1) a group of people standing around a luggage cart.\newline 2) a group of people standing next to a luggage cart.  &  0) a group of people standing around a luggage carousel.\newline 1) a group of people standing around a luggage cart.\newline 2) a group of people standing next to each other.  \\  \hline

\raisebox{-1.218\height}{\includegraphics[width=0.52in]{COCO_train2014_000000521184.jpg}}  &  horse wrangler(s), Tennessee walker, wild horse  &  \textbf{found on ranch}  &  0) a man standing next to a brown horse.\newline 1) a man is standing next to a horse\newline 2) a man standing next a brown horse in a \textbf{stable}.  &  0) a couple of people standing next to a horse.\newline 1) a woman standing next to a brown horse.\newline 2) a couple of people standing next to a brown horse.  &  0) a group of people standing next to a horse.\newline 1) a group of people standing next to a brown horse.\newline 2) a group of men standing next to a brown horse.  \\  \hline
\end{tabular}
}
\end{center}
\end{table*}
\subsection{Performance Comparison}
We compare the performance of CNet-NIC with that of several state-of-the-art image captioning methods (as reported in the respective papers):

\begin{itemize}
\item {\bf Neural Image Caption (NIC)}\cite{vinyals2015show}, which uses a vector space representation of image features produced by a CNN to initialize an LSTM-based RNN trained to generate image captions from vector space representation of image features.  

\item {\bf Hard and Soft Attention}\cite{xu2015show}, which combines two attention-based image captioning mechanisms under an encoder-decoder framework: a “soft” deterministic attention mechanism trainable by standard back-propagation methods and 2) a “hard” stochastic attention mechanism that is trained using reinforcement learning.

\item {\bf LRCN}\cite{donahue2015long} which combines CNN with LSTMs to perform visual recognition and image captioning. 

\item {\bf ATT}\cite{you2016image}  which combines top-down and bottom-up attention models to extract image features that are used to train an RNN to produce image captions.

\item {\bf Sentence-Condition}\cite{DBLP:journals/corr/ZhouXKC16} which uses a  \textit{text-conditional attention} mechanism for focusing the caption generator  on specific image features that should inform the caption given the already generated caption text.  

\item {\bf LSTM-A}\cite{yao2017boosting} which extends the basic  LSTM model with image attributes model by rearranging image and attributes input in different positions and time to boost the accuracy of image captioning. 

\end{itemize}

Table \ref{table_performancecomparison} shows the performance of each method on MS COCO image captioning data set. Bold represents the best in that metric and italic represents the second best. Overall the performance of CNet-NIC  is comparable to or better than all other models on all measures, especially with respect to CIDEr-D, the only measure that is explicitly designed for the purpose of evaluating image captions.

\subsection{CNet-NIC Ablation Study Results}
We report results of an {\em ablation study} of CNet-NIC, where we examine the relative contributions of the different components of the CNet-NIC architecture. 

From the results summarized in Table \ref{table_factoranalysis}, we  see that detected objects and directly related terms contribute to greater improvements in performance as compared to indirectly related terms.  We conjecture that the detected objects and directly related terms provide more information about the individual objects in an image whereas the indirectly related terms provide information about the scene as a whole. This perhaps explains why only adding indirectly related terms to image embedding improves performance as measured by METEOR, ROUGE-L and CIDEr-D, albeit at the cost of a slight decrease in BLEU. We further note that the indirectly related terms contribute to increases in CIDEr-D, even when no image features are available. Overall, we find that CNet-NIC which combines the background knowledge (ConceptNet derived terms) related to the detected objects and the scene in generating image captions outperforms all other methods that do not make use of such background knowledge.

\subsection{Qualitative  Analysis of Captions}

Table \ref{table_invitingcommonsense} presents several representative examples of captions produced by CNet-NIC. Here we take a qualitative  look at the captions to explore the role played by the commonsense or background knowledge provided by the ConceptNet knowledge graph.  In the first  example, the ConceptNet derived terms such as "upholstered", "found in house", etc.  appear to yield more accurate captions.   In the third  example, the standard model and the model without indirectly related terms completely ignore  the large furniture such as tables and chairs while the model that incorporates indirectly related terms such as "item of furniture", "reupholstery", "end table", etc. leads to what appear to be  better captions. For the fourth example, the indirectly related terms appear to yield a more accurate caption model, e.g., one that mentions the book rack.   For the sixth image listed, only the model with indirectly related terms as "dairy farm", "feed lot", etc. from the knowledge graph correctly recognizes that the scene is occurring in a "barn". In the seventh example, the model with indirectly related terms correctly deduces that most people in the image are travelers and conclude that they are in a baggage claim area. These examples offer further qualitative evidence that shows the utility and effectiveness of background knowledge supplied by knowledge graphs to improve the quality of image captions.

\section{Summary and Discussion}

The focus of this paper is on  the image captioning problem, which requires analyzing the visual content of an image, and generating a caption, i.e., a textual description that summarizes the most salient aspects of the image. Image captioning presents several challenges beyond those addressed by object recognition, e.g., inferring information that is not explicitly depicted in the image. However, existing methods for image captioning (See  \cite{bernardi2016automatic} for a review) fail to take advantage of readily available general or commonsense knowledge about the world.   

In this paper, we have presented CNet-NIC, an approach
to image captioning that incorporates background knowledge
available in the form of knowledge graphs to augment
the information extracted from images.  The results of our experiments on several benchmark data sets, such as MS COCO,  show that the variants of the state-of-the-art methods for image captioning that make use of the information extracted from knowledge graphs, e.g., CNet-NIC, can substantially
outperform those that rely solely on the information
extracted from images, e.g., CNet.
 
Some promising directions for future work include: variants and extensions of CNet-NIC, including those that substantially improve the quality of captions, provide justifications for the captions that they produce, tailor captions for visual question answering, tailoring captions to different audiences and contexts, etc. by bringing to bear on such tasks, all available background knowledge.

\section*{Acknowledgements}
This project was supported in part by the National Center for Advancing Translational Sciences, National Institutes of Health through the grant UL1 TR000127 and TR002014, by the National Science Foundation, through the grants 1518732, 1640834, and 1636795, the Pennsylvania State University’s Institute for Cyberscience and the Center for Big Data Analytics and Discovery Informatics, the Edward Frymoyer Endowed Professorship in Information Sciences and Technology at Pennsylvania State University and the Sudha Murty Distinguished Visiting Chair in Neurocomputing and Data Science funded by the Pratiksha Trust at the Indian Institute of Science [both held by Vasant Honavar]. The content is solely the responsibility of the authors and does not necessarily represent the official views of the sponsors.

\newpage
\clearpage

\newpage
\clearpage


\end{document}